\documentclass[letterpaper]{article} 
\usepackage{aaai2026}  
\usepackage{times}  
\usepackage{helvet}  
\usepackage{courier}  
\usepackage[hyphens]{url}  
\usepackage{graphicx} 
\usepackage{natbib}  
\usepackage{caption} 
\usepackage{algorithm}
\usepackage{algorithmic}

\usepackage{newfloat}
\usepackage{listings}
\DeclareCaptionStyle{ruled}{labelfont=normalfont,labelsep=colon,strut=off} 
\floatstyle{ruled}
\newfloat{listing}{tb}{lst}{}
\floatname{listing}{Listing}
\usepackage{amssymb}
\usepackage{amsmath}
\usepackage{booktabs}
\usepackage{multirow}
\usepackage[table,svgnames]{xcolor}
\definecolor{mydeepgreen}{RGB}{0,180,0}
\usepackage{soul}
\usepackage{subcaption}  
\usepackage{bm}

\title{VasoMIM: Vascular Anatomy-Aware Masked Image Modeling \\ for Vessel Segmentation}
\author{
    De-Xing Huang$^{1,2}$,
    Xiao-Hu Zhou$^{1,2*}$,
    Mei-Jiang Gui$^{1,2}$,
    Xiao-Liang Xie$^{1,2}$,
    Shi-Qi Liu$^{1}$, \\
    Shuang-Yi Wang$^{1,2}$,
    Tian-Yu Xiang$^{1,2}$,
    Rui-Ze Ma$^{1}$,
    Nu-Fang Xiao$^{1}$,
    Zeng-Guang Hou$^{1,2}$\thanks{Corresponding Authors.}
}

\affiliations{
    $^1$State Key Laboratory of Multimodal Artificial Intelligence Systems, Institute of Automation, Chinese Academy of Sciences \\
    $^2$School of Artificial Intelligence, University of Chinese Academy of Sciences \\
    Emails: \{huangdexing2022, xiaohu.zhou, zengguang.hou\}@ia.ac.cn
}

\usepackage{bibentry}

\begin{document}

\maketitle

\begin{abstract}
Accurate vessel segmentation in X-ray angiograms is crucial for numerous clinical applications. However, the scarcity of annotated data presents a significant challenge, which has driven the adoption of self-supervised learning (SSL) methods such as masked image modeling (MIM) to leverage large-scale unlabeled data for learning transferable representations. Unfortunately, conventional MIM often fails to capture vascular anatomy because of the severe class imbalance between vessel and background pixels, leading to weak vascular representations. To address this, we introduce \underline{\textbf{Vas}}cular anat\underline{\textbf{o}}my-aware \underline{\textbf{M}}asked \underline{\textbf{I}}mage \underline{\textbf{M}}odeling (\textbf{VasoMIM}), a novel MIM framework tailored for X-ray angiograms that explicitly integrates anatomical knowledge into the pre-training process. Specifically, it comprises two complementary components: \textit{anatomy-guided masking strategy} and \textit{anatomical consistency loss}. The former preferentially masks vessel-containing patches to focus the model on reconstructing vessel-relevant regions. The latter enforces consistency in vascular semantics between the original and reconstructed images, thereby improving the discriminability of vascular representations. Empirically, VasoMIM achieves state-of-the-art performance across three datasets. These findings highlight its potential to facilitate X-ray angiogram analysis.
\end{abstract}

\begin{links}
    \link{Project Page}{https://dxhuang-casia.github.io/VasoMIM}
    \link{Extended Version}{https://arxiv.org/abs/2508.10794}
\end{links}


\section{Introduction}
Cardiovascular diseases (CVDs) constitute a global health crisis and remain the leading cause of mortality worldwide~\cite{vaduganathan2022global}. X-ray angiography is considered the gold standard for diagnosing CVDs~\cite{kheiri2022computed}, planning treatment~\cite{writing20222021}, and guiding intraoperative procedures~\cite{huang2025real}. However, radiologists often struggle to accurately delineate vessels in X-ray angiograms because of low contrast, motion artifacts, and overlapping anatomical structures~\cite{huang2024spironet}. Consequently, there is an urgent need for automated vessel segmentation methods.

\begin{figure}[htbp]
    \centerline{\includegraphics{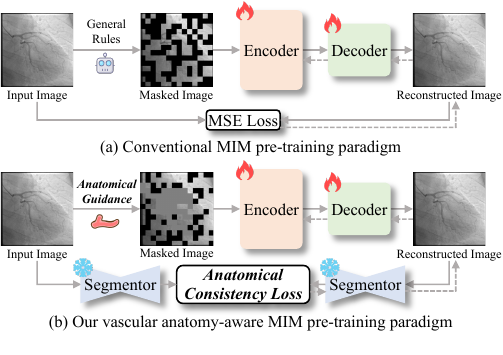}}
	\caption{Comparison of conventional MIM and VasoMIM. (a) Conventional MIM masks patches based on \textit{general rules} and learns to reconstruct patches via minimizing \textit{pixel-level loss}. (b) VasoMIM guides patch masking with \textit{vascular anatomy} and enforces \textit{anatomical consistency} during reconstruction, enabling the model to learn richer vascular representations. Dark gray patches are vessel-relevant regions.
    }
	\label{fig:motivation}
\end{figure}

Over the past decade, numerous vessel segmentation algorithms have been proposed~\cite{ronneberger2015u,huang2024spironet,wu2025denver}, helping to alleviate radiologists’ workload. However, training high-performance models requires large-scale datasets of X-ray angiograms with pixel-level labels, and producing such annotations remains labor-intensive, time-consuming, and dependent on specialized domain knowledge~\cite{esteva2019guide}. Self-supervised learning (SSL) provides an attractive alternative by learning generalizable representations from vast unlabeled data, thereby boosting downstream performance~\cite{gui2024survey}. In particular, masked image modeling (MIM)~\cite{he2022masked,fu2025rethinking,zhuang2025mim,tang2025mambamim} has achieved remarkable success in natural and medical image analysis by training models to reconstruct masked image patches, as shown in Fig.~\ref{fig:motivation}~(a).

Nevertheless, adapting MIM to X-ray angiograms remains challenging due to the \textit{extreme class imbalance} between vessel and background pixels. We attribute this difficulty to a lack of explicit anatomical awareness in two key aspects of the MIM framework. \textbf{First, vessel-containing patches are more likely to be ignored than background-only patches during the masking process.} Current masking strategies are typically based on general rules, which can be divided into data-independent and data-adaptive~\cite{hinojosa2024colormae}. The former includes random masking~\cite{he2022masked}, block-wise masking~\cite{bao2022beit}, uniform masking~\cite{li2022uniform}, \textit{etc}. The latter involves designing masking strategies based on specific feedback, \textit{e.g.,} attention maps~\cite{kakogeorgiou2022hide,liu2023good}, loss predictions~\cite{wang2025bootstrap}, reward functions~\cite{xu2025self}, \textit{etc}. However, none of these methods adequately focus on vascular anatomy, resulting in few vessel-containing patches being masked and hindering the effective learning of vascular representations. \textbf{Second, the pixel-level reconstruction loss fails to preserve semantic consistency during reconstruction.} Most MIM methods minimize the mean squared error (MSE) between original and reconstructed patches, but this pixel-level objective ignores vascular anatomy and thus fails to encourage learning of discriminative vascular representations.

To address these challenges, we introduce \underline{\textbf{Vas}}cular anat\underline{\textbf{o}}my-aware \underline{\textbf{M}}asked \underline{\textbf{I}}mage \underline{\textbf{M}}odeling (\textbf{VasoMIM}), a novel MIM paradigm tailored for X-ray angiograms, as presented in Fig.~\ref{fig:motivation} (b). The core insight of VasoMIM is to inject vascular anatomical knowledge into MIM. \textbf{To address the first challenge, we introduce an anatomy-guided masking strategy} that biases the masking process toward vessel-containing patches, guiding the model to reconstruct anatomically relevant regions. \textbf{For the second challenge, we propose an anatomical consistency loss} that enforces consistency between the vascular anatomy from the original and the reconstructed angiograms, thereby encouraging the model to learn more discriminative vascular representations. This raises the question of how to extract vascular anatomy from X-ray angiograms. To resolve this, we apply Frangi filter \cite{frangi1998multiscale} to obtain vessel segmentation masks in an unsupervised manner. To integrate this filter into end-to-end SSL, we train a segmentor in advance using pseudo-labels produced by it.

In summary, our main contributions are as follows:
\begin{itemize}
    \item A novel vascular anatomy-aware MIM framework, VasoMIM, is proposed to enhance the model's ability to understand vascular content in X-ray angiograms.
    \item Two complementary components are proposed: (1) an anatomy-guided masking strategy that preferentially masks vessel-containing patches, guiding the model to focus on vascular regions, and (2) an anatomical consistency loss that ensures semantic consistency between the original and reconstructed images, thereby boosting the discriminability of vascular representations.
    \item Extensive experiments demonstrate the benefits of integrating anatomical knowledge into the MIM framework. Meanwhile, VasoMIM consistently outperforms state-of-the-art SSL alternatives on vessel segmentation tasks.
\end{itemize}

\section{Related Work}
\subsection{SSL in Medical Imaging}
Self-supervised learning (SSL)~\cite{chen2020simple, chen2021empirical, caron2021emerging, he2022masked, wang2025bootstrap} offers a practical solution to mitigate annotation scarcity in medical imaging. Existing approaches can be categorized into two main paradigms: contrastive learning (CL)-based and masked image modeling (MIM)-based. CL-based methods aim to pull positive pairs together while pushing negative pairs apart in feature space, such as C$2$L~\cite{zhou2020comparing}, MICLe~\cite{azizi2021big}, VoCo~\cite{wu2024voco}, RAD-DINO~\cite{perez2025exploring}, \textit{etc}. However, because these methods focus on image-level representations, they may perform poorly on dense prediction tasks like segmentation~\cite{chaitanya2020contrastive}. MIM-based approaches train models to reconstruct masked patches, yielding finer-grained representations~\cite{yuan2023hap}. The choice of training objective is critical. Most methods use a pixel-level reconstruction loss~\cite{kang2024deblurring, li2024anatomask, xu2025self}, while some incorporate a contrastive loss to improve multi-view alignment in 3D medical images~\cite{zhuang2025advancing, zhuang2025mim}. However, these objectives do not ensure semantic consistency in the reconstructed images, often resulting in weak vascular representations. Our work addresses this limitation by introducing a semantic-level anatomical consistency loss.

\subsection{Masking Strategies in MIM}
The design of the masking strategy is crucial in MIM given the high redundancy of images~\cite{he2022masked}. Most methods~\cite{xie2024rethinking,kang2024deblurring} adopt random masking at high ratios (\textit{e.g.,} $75\%$). To create more challenging pretext tasks, many carefully designed masking strategies have been proposed. AMT~\cite{liu2023good} masked patches based on attention maps. SemMAE~\cite{li2022semmae} learned semantic parts of images first and used part segmentation results to guide patch masking. HAP~\cite{yuan2023hap} exploited human-structure priors to guide the masking process for human-centric perception pre-training. Methods such as HPM~\cite{wang2025bootstrap}, AnatoMask~\cite{li2024anatomask} and, AHM~\cite{xu2025self} identified patches that are difficult to reconstruct through a loss predictor, a self-distillation framework and a policy network, respectively, and preferentially masked these hard patches. However, none of these approaches can explicitly incorporate vascular anatomy into the masking process, which is essential for guiding models to focus on vascular regions.

\begin{figure*}[t]
	\centering
	\centerline{\includegraphics{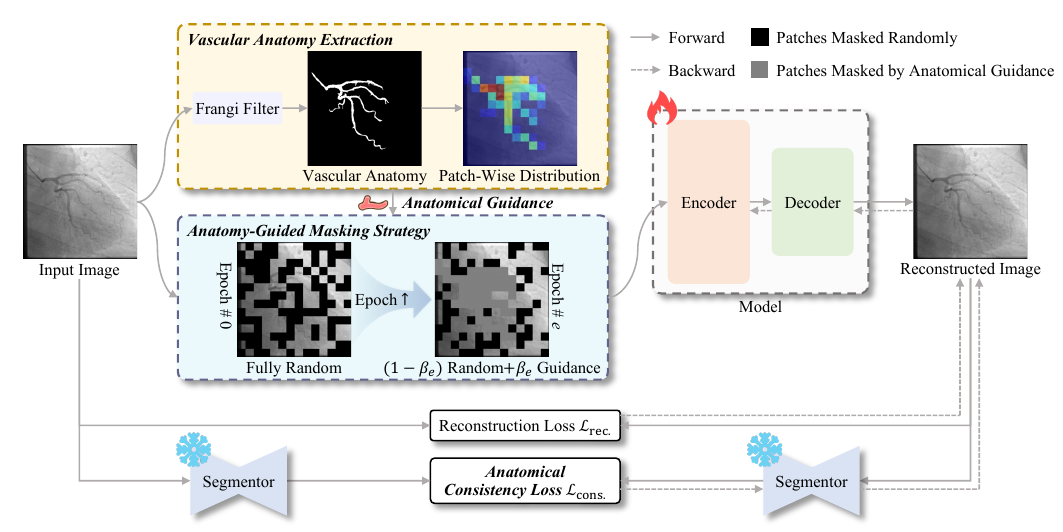}}
	\caption{Overall framework of VasoMIM. During pre-training, each X-ray angiogram is first processed by Frangi filter to extract its vascular anatomy. From this anatomy, we derive a patch-wise vascular anatomical distribution $f$ to guide the masking process. Finally, the model is optimized by minimizing $\mathcal{L}_{\rm train}$, which is a combination of standard pixel-wise reconstruction loss $\mathcal{L}_{\rm rec.}$ and the designed anatomical consistency loss $\mathcal{L}_{\rm cons.}$.}
	\label{fig:framework}
\end{figure*}

\subsection{Vessel Segmentation}
Traditional vessel segmentation methods, such as Frangi filter~\cite{frangi1998multiscale}, active contours~\cite{taghizadeh2014local}, and graph cuts~\cite{wang2020tensor}, rely on handcrafted features that generalize poorly across the wide variability of X-ray angiograms. The emergence of deep learning has significantly advanced the field. Early CNN-based architectures like U-Net~\cite{ronneberger2015u} and its variants~\cite{zhou2019unet++,li2020cau} capture hierarchical features but remain constrained by their local receptive fields. More recent transformer- and Mamba-based models~\cite{chen2024transunet, hatamizadeh2021swin, ruan2024vm, wang2024lkm} address this limitation by integrating global context, thereby improving the delineation of complex vascular structures. However, these deep models are highly data-hungry, and the scarcity of labeled angiograms continues to limit their gains~\cite{zhou2021review}. This work leverages MIM to learn generalizable representations from large-scale unlabeled data, significantly boosting the performance of segmentors. We use U-Net as a representative case study to demonstrate the broad applicability of our method.

\section{Method}
The overall framework is illustrated in Fig.~\ref{fig:framework}. First, we extract vascular anatomy from X-ray angiograms using Frangi filter~\cite{frangi1998multiscale}. Next, this anatomical knowledge is used to guide the masking process. Finally, an anatomical consistency loss is incorporated into the training objective to capture discriminative vascular representations.

\subsection{Vascular Anatomy Extraction}
Frangi filter effectively enhances vascular anatomy in an unsupervised manner by highlighting vessel-like features. Following the implementation in \cite{wu2025denver}, our approach proceeds in three stages.

\noindent \textbf{Multi-Scale Hessian Vesselness.} Given an X-ray angiogram $I\in \mathbb{R}^{C\times H\times W}$, we smooth it with a Gaussian kernel $G\left(\sigma\right)$ at scales $\sigma\in\left\{1, 2, 3, 4\right\}$ and calculate the multi-scale Hessian $H_\sigma(i) =\nabla^2\left[I * G(\sigma)\right]\left(i\right)$, where $C$, $H$, and $W$ denote the number of channels, height, and width of the image. At each pixel $i$, we obtain eigenvalues $|\lambda_1| \leq |\lambda_2|$ of $H_{\sigma}\left(i\right)$ and define the vesselness response
\begin{align}
V_{\sigma}(i) =
\begin{cases}
\lvert \lambda_{1}(i)\rvert, & \lambda_{2}(i) < 0,\\
0, & {\rm otherwise.}
\end{cases}
\end{align}

The final vesselness map $V\left(i\right)$ is the maximum response over all scales, capturing vessels of varying diameters.

\noindent \textbf{Adaptive Percentile Thresholding.} Subsequently, we apply adaptive thresholding to the vesselness map $V$ at its $\alpha$-th percentile value $T={\rm Percentile}_{\alpha}\left[V\left(i\right)\right]$, creating a binary map $\hat{B}\left(i\right)=\mathbb{I}\left[V(i)\geq T\right]$. This approach effectively filters out low-intensity noise while retaining prominent vascular anatomy.

\noindent \textbf{Seed Region Growing.} Finally, an optimal seed pixel $s=\arg\max_i V(i)$ is automatically identified based on the highest vesselness intensity value. A region-growing algorithm~\cite{adams1994seeded} is then applied to the binary threshold map $\hat{B}$, iteratively expanding the region to include neighboring pixels with high vesselness values. This process yields a binary mask $B\in\mathbb{R}^{1\times H\times W}$ that accurately delineates the vascular anatomy.

\subsection{Anatomy-Guided Masking Strategy}
Compared to natural images, X-ray angiograms exhibit far greater spatial redundancy because vessels occupy only a small fraction of each image. Existing general rule-based masking strategies lack any anatomical awareness and therefore mask background-only patches far more often. As a result, pre-training is dominated by background content rather than learning vascular representations.

To address this issue, we introduce an anatomy-guided masking strategy. Our key idea is that patches containing vessels are more informative and should be masked with higher probability. Formally, a patch-wise vascular anatomical distribution $f$ is defined. In the pre-training process, X-ray angiograms $I$ and the corresponding masks $B$ are split into non-overlapping patches $x\in\mathbb{R}^{N\times\left(P^2C\right)}$ and $m\in\mathbb{R}^{N\times P^2}$, respectively, where $P$ is the patch size and $N=HW/P^2$ is the sequence length. Let $m_i\in \{0, 1\}^{P^2}$ be the $i$-th patch of $m$. Then $f(m_i)$ is defined as follows:
\begin{align}
f\left(m_i\right)=\frac{\sum_{j=1}^{P^2}\mathbb{I}\left(m_{ij}=1\right)}{\sum_{i,j=1}^{N,P^2}\mathbb{I}\left(m_{ij}=1\right)},\quad i=1, 2, \cdots, N
\end{align}
where $\mathbb{I}\left(\cdot\right)$ is the indicator function.

\noindent \textbf{Weak-to-Strong Anatomical Guidance.} By leveraging the patch-wise vascular anatomical distribution, we perform anatomy-aware patch masking. However, during the early training stages, masking too many vessel-containing patches may impair the model’s ability to reconstruct those patches rich in vascular content. To this end, we adopt a weak-to-strong anatomy-guided strategy. As illustrated in Fig.~\ref{fig:framework}, for a specific training epoch $e$, $\beta_e$ of the masked patches are sampled according to $f$, and the remaining $1-\beta_e$ are randomly selected. $\beta_e$ is increased linearly during pre-training.
\begin{align}
    \beta_e=\beta_0+\frac{e}{E}\left(\beta_E-\beta_0\right)
\end{align}
where $E$ is the maximum pre-training epoch, and $\beta_0, \beta_E\in [0, 1]$ are hyper-parameters. Under this masking strategy, $\beta_e\gamma N$ patches are masked by anatomical guidance and the remaining $(1-\beta_e)\gamma N$ patches are randomly masked, where $\gamma$ represents the masking ratio.

\subsection{Anatomical Consistency Loss}
In conventional MIM, the reconstruction objective typically minimizes the mean squared error (MSE) between masked and reconstructed patches. However, this loss fails to preserve semantic consistency and does not encourage the model to learn the discriminative vascular representations essential for downstream tasks.

To overcome this limitation, we introduce an anatomical consistency loss that explicitly directs the model to preserve vascular anatomy during reconstruction:
\begin{table}[htbp]
\centering
\renewcommand\arraystretch{1.2}
\begin{tabular}{llll}
\toprule
Usage & Dataset & \# Train & \# Test \\ \midrule
\multirow{6}{*}{Pre-Training} & ARCADE & $2,000$ & $-$ \\
 & CADICA & $6,594$ & $-$ \\
 & Stenosis & $7,492$ & $-$ \\
 & SYNTAX & $2,943$ & $-$ \\
 & XCAD & $1,621$ & $-$ \\
 & \cellcolor{gray!20}Total & \cellcolor{gray!20}$20,650$ & \cellcolor{gray!20}$-$ \\
\midrule
\multirow{4}{*}{Vessel Segmentation} & ARCADE & $200$ & $300$ \\
 & XCAV & $175$ & $46$ \\
 & CAXF & $337$ & $201$\\ 
 & \cellcolor{gray!20}Total & \cellcolor{gray!20}$712$ & \cellcolor{gray!20}$547$ \\
\bottomrule
\end{tabular}
\caption{
Details of datasets used for pre-training and vessel segmentation.
}\label{table:datasets}

\end{table}
\begin{align}
    \mathcal{L}_{\rm cons.}=\mathcal{L}\left[\mathcal{S}\left(I\right),\mathcal{S}\left(I^\prime\right)\right]
\end{align}
where $I^\prime$ is the reconstructed X-ray angiogram. $\mathcal{L}\left(\cdot, \cdot\right)$ is an abstract metric function, and we use cross-entropy loss by default. Here, $\mathcal{S}\left(\cdot\right)$ represents the segmentor used to extract vascular anatomy. Although Frangi filter provides high-quality vascular masks, its non-differentiability prevents direct integration into the end-to-end pre-training process. To solve this problem, we train a lightweight UNeXt-S network ({$\sim0.3$} M)~\cite{valanarasu2022unext} on pseudo-labels generated by Frangi filter, then freeze its weights during subsequent pre-training.

\textbf{Training Objective.} In addition to the anatomical consistency loss, we include the pixel-level reconstruction loss (\textit{i.e.,} MSE loss) following conventional MIM approaches. The overall training loss is given by:
\begin{align}
    \mathcal{L}_{\rm train} = \mathcal{L}_{\rm rec.} + \mathcal{L}_{\rm cons.}
\end{align}

\section{Results}
\subsection{Datasets}
Table~\ref{table:datasets} summarizes the datasets used for both pre-training and vessel segmentation.

\noindent\textbf{Pre-Training.} We assemble a corpus of $20,650$ coronary X-ray angiograms from five public datasets: ARCADE~\cite{popov2024dataset}, CADICA~\cite{jimenez2024cadica}, Stenosis~\cite{danilov2021real}, SYNTAX~\cite{mahmoudi2025x}, and XCAD~\cite{ma2021self}.

\noindent\textbf{Vessel Segmentation.} VasoMIM is evaluated on three benchmarks: one in-domain dataset, ARCADE, and two out-of-domain datasets, CAXF~\cite{li2020cau} and XCAV~\cite{wu2025denver}. Note that the images of ARCADE used for vessel segmentation are not included in the pre-training set.

\begin{table*}[t]
\centering
\renewcommand\arraystretch{1.2}

\begin{tabular}{lcccccc}
\toprule
\multirow{2}{*}{Method} & \multicolumn{2}{c}{ARCADE} & \multicolumn{2}{c}{CAXF} & \multicolumn{2}{c}{XCAV} \\
 & DSC (\%) & clDice (\%) & DSC (\%) & clDice (\%) & DSC (\%) & clDice (\%) \\ \midrule
\multicolumn{7}{l}{\textit{Traditional}}                                                       \\
 Frangi Filter~\cite{frangi1998multiscale} & $41.30$ & $40.91$ & $64.01$ & $65.73$ & $58.46$ & $57.15$ \\
 \midrule
\multicolumn{7}{l}{\textit{From Scratch}}                                                            \\
 U-Net~\cite{ronneberger2015u} & $58.27${\tiny$\pm1.33$} & $59.70${\tiny$\pm1.40$} & $78.72${\tiny$\pm0.74$} & $82.68${\tiny$\pm0.87$} & $68.63${\tiny$\pm2.80$} & $63.47${\tiny$\pm3.33$} \\
\midrule
\multicolumn{7}{l}{\textit{Contrastive Learning}}                                                    \\
 MoCo v3~\cite{chen2021empirical} & $60.99${\tiny$\pm0.30$} & $62.68${\tiny$\pm0.18$} & $77.76${\tiny$\pm0.51$} & $80.91${\tiny$\pm0.31$} & $70.85${\tiny$\pm0.34$} & $63.97${\tiny$\pm0.71$} \\
 DINO~\cite{caron2021emerging} & $65.86${\tiny$\pm0.49$} & $67.84${\tiny$\pm0.52$} & $80.13${\tiny$\pm0.53$} & $82.90${\tiny$\pm0.51$} & $72.28${\tiny$\pm0.96$} & $66.36${\tiny$\pm1.17$} \\
\midrule
\multicolumn{7}{l}{\textit{Masked Image Modeling}}                                                   \\
 MAE~\cite{he2022masked} & $68.17${\tiny$\pm0.29$} & $69.89${\tiny$\pm0.22$} & $83.53${\tiny$\pm0.14$} & $87.37${\tiny$\pm0.21$} & $76.43${\tiny$\pm0.17$} & $72.58${\tiny$\pm0.49$} \\
 SimMIM~\cite{xie2022simmim} & $66.92${\tiny$\pm0.43$} & $68.93${\tiny$\pm0.71$} & $82.24${\tiny$\pm0.34$} & $85.77${\tiny$\pm0.17$} & $75.10${\tiny$\pm0.36$} & $69.98${\tiny$\pm0.42$} \\
 AMT~\cite{liu2023good} & $68.15${\tiny$\pm0.23$} & $69.77${\tiny$\pm0.38$} & $83.47${\tiny$\pm0.09$} & $87.40${\tiny$\pm0.04$} & $76.51${\tiny$\pm0.20$} & $72.60${\tiny$\pm0.44$} \\
 DeblurringMIM$^\dag$~\cite{kang2024deblurring} & $\underline{68.60}${\tiny$\pm0.44$} & $70.21${\tiny$\pm0.37$} & $\underline{83.85}${\tiny$\pm0.09$} & $\underline{87.78}${\tiny$\pm0.20$} & $\underline{77.02}${\tiny$\pm0.08$} & $\underline{73.58}${\tiny$\pm0.19$} \\
 CrossMAE~\cite{fu2025rethinking} & $62.40${\tiny$\pm0.33$} & $64.23${\tiny$\pm0.27$} & $80.07${\tiny$\pm0.13$} & $83.45${\tiny$\pm0.19$} & $72.25${\tiny$\pm0.24$} & $65.94${\tiny$\pm0.15$} \\
 HPM~\cite{wang2025bootstrap} & $66.82${\tiny$\pm0.28$} & $68.49${\tiny$\pm0.41$} & $82.61${\tiny$\pm0.21$} & $86.18${\tiny$\pm0.10$} & $75.48${\tiny$\pm0.19$} & $70.79${\tiny$\pm0.26$} \\
 CheXWorld$^\dag$~\cite{yue2025chexworld} & $67.95${\tiny$\pm0.26$} & $\underline{70.31}${\tiny$\pm0.48$} & $80.64${\tiny$\pm0.31$} & $82.65${\tiny$\pm0.31$} & $73.74${\tiny$\pm0.24$} & $67.13${\tiny$\pm0.32$} \\
 \textbf{VasoMIM} & $\bm{68.85}${\tiny$\pm0.47$} & $\bm{70.56}${\tiny$\pm0.36$} & $\bm{84.49}${\tiny$\pm0.17$} & $\bm{88.33}${\tiny$\pm0.09$} & $\bm{77.52}${\tiny$\pm0.26$} & $\bm{74.18}${\tiny$\pm0.34$} \\
\bottomrule
\end{tabular}
\caption{Comparison of state-of-the-art methods on ARCADE, CAXF and XCAV. All methods are reimplemented using their official codebases. The best results are highlighted in \textbf{bold} and the second-best results are \underline{underlined}. Results are reported as ``${\rm mean}\pm{\rm std}$'' over three random seeds, except for Frangi filter. $^\dag$ indicates that the model is specialized in medical imaging.
}
\label{table:sota}
\end{table*}

\subsection{Implementation Details}
\textbf{Pre-Training.} Our implementation is based on MAE~\cite{he2022masked}. By default, ViT-B/16~\cite{dosovitskiy2021image} is used as the backbone following previous works. VasoMIM is pre-trained for $800$ epochs on an NVIDIA ${\rm A}6000$ GPU, employing the AdamW optimizer~\cite{loshchilov2019decoupled} with a batch size of $256$ and an input resolution of $224\times224$.

\noindent \textbf{Vessel Segmentation.} We adopt U-Net~\cite{ronneberger2015u} as the segmentation decoder and fine-tune it end-to-end on ARCADE, CAXF and XCAV for $500$ epochs each, using input images resized to $224\times224$. Optimization is performed using AdamW with an initial learning rate of $1e^{-4}$ and a weight decay of $0.05$. A cosine-annealing schedule with $\text{$T$\_{max} = $500$}$ is employed. All experiments are conducted on an NVIDIA ${\rm A}6000$ GPU.

\noindent \textbf{Evaluation Metrics.} Dice similarity coefficient (DSC) and centerlineDice (clDice)~\cite{shit2021cldice} are adopted. Compared to DSC, clDice better captures topological correctness by measuring overlap between predicted and ground-truth vascular centerlines.

\subsection{Main Results on Vessel Segmentation}
All baselines are pre-trained on the angiogram dataset in Table~\ref{table:datasets} using their official implementations. Each model is fine-tuned with three random seeds, and we report all metrics as ``${\rm mean}\pm{\rm std}$".

\noindent \textbf{In-Domain Dataset (ARCADE).} When trained from scratch, U-Net achieves $58.27\%$ DSC and $59.70\%$ clDice on ARCADE. By pre-training on large-scale unlabeled data, our VasoMIM improves performance by $10.58\%$ in DSC and $10.86\%$ in clDice, reaching $68.85\%$ DSC and $70.56\%$ clDice, respectively, surpassing leading SSL baselines by a clear margin. Compared to Frangi filter, VasoMIM yields an absolute gain of $27.55\%$ in DSC and $29.65\%$ in clDice, underscoring the crucial role of our anatomy-aware pre-training in boosting segmentation performance.

\begin{table}[htbp]
\centering
\renewcommand\arraystretch{1.2}
\begin{tabular}{llll}
\toprule
Guidance & $\mathcal{L}_{\rm cons.}$ & ARCADE & CAXF \\ \midrule
 $-$ & $-$ & $68.00$ & $83.15$ \\ 
 $-$ & \checkmark & $68.45$ & $84.03$ \\
 \checkmark & $-$ & $68.30$ & $83.96$ \\
 \cellcolor{gray!20}\checkmark & \cellcolor{gray!20}\checkmark & \cellcolor{gray!20}$\mathbf{68.85}$ & \cellcolor{gray!20}$\mathbf{84.49}$ \\
\bottomrule
\end{tabular}
\caption{
Ablation study of the proposed anatomy-guided masking strategy and anatomical consistency loss. DSC is reported in this table.
}
\label{table:ablation}
\end{table}


\begin{figure*}[htbp]
    \centering
    \begin{subfigure}[b]{0.48\textwidth}
        \centering
        \includegraphics{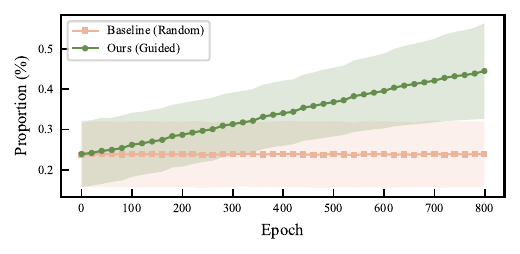}
        \caption{}
    \end{subfigure}
    \hfill
    \begin{subfigure}[b]{0.48\textwidth}
        \centering
        \includegraphics{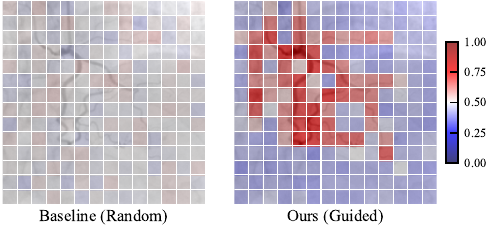}
        \caption{}
    \end{subfigure}
    \caption{Some evidence of anatomy-guided masking strategy. (a) Proportion of vessel-containing patches in the masked patches during pre-training. (b) Patch-wise masking ratio over the pre-training process, \textit{i.e.}, $\frac{1}{E}\sum_{j=1}^E\mathbb{I}\left(\text{Patch $x_i$ is masked in epoch $j$}\right)$.}
    \label{fig:vessel_patch_ratio}
\end{figure*}

\noindent \textbf{Out-of-Domain Datasets (CAXF and XCAV).} We further conduct experiments on out-of-domain datasets. Our VasoMIM consistently achieves the best performance, with $84.49\%$ and $77.52\%$ DSC on CAXF and XCAV, respectively. Notably, it outperforms the best baseline by $+0.64\%$ DSC on CAXF and $+0.50\%$ DSC on XCAV. This performance gap is even larger than that on the in-domain dataset (\textit{i.e.}, $+0.25\%$ DSC), highlighting VasoMIM’s strong generalizability and robustness.

\begin{figure}[htbp]
	\centerline{\includegraphics{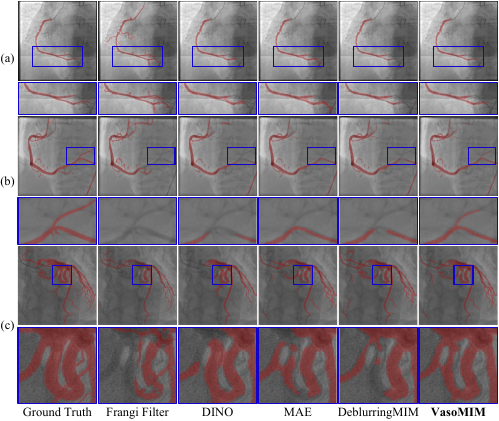}}
	\caption{Qualitative results on (a) ARCADE, (b) CAXF, and (c) XCAV. Details are zoomed in within blue boxes.}
	\label{fig:SOTA_Vis}
\end{figure}

\noindent \textbf{Qualitative Results.} Several cases from the three datasets are presented in Fig.~\ref{fig:SOTA_Vis}. VasoMIM produces precise vessel segmentations, even for very thin or blurred vessels.

\subsection{Ablation Study}
Ablation studies are conducted on ARCADE and CAXF with default settings highlighted in \sethlcolor{gray!20}\hl{gray}. Each model is fine-tuned with three random seeds. Table~\ref{table:ablation} shows the results of sequentially adding each proposed component to the vanilla baseline, \textit{i.e.}, MAE ($\gamma=0.5$).

\noindent \textbf{Effectiveness of Anatomy-Guided Masking Strategy.} Simply using the anatomy-guided masking strategy yields a clear performance boost ($+0.30\%$ DSC on ARCADE and $+0.81\%$ DSC on CAXF). To better understand this improvement, we measure the proportion of vessel-containing patches in the masked patches during the pre-training process. As shown in Fig.~\ref{fig:vessel_patch_ratio}(a), our strategy masks a significantly larger number of vessel-containing patches, whereas the baseline masks a small and relatively constant proportion. We further visualize an example in Fig.~\ref{fig:vessel_patch_ratio} (b), where the masking ratio of each patch over pre-training is colored. Our strategy clearly favors masking patches rich in vascular anatomy, whereas the baseline shows no such preference.

\noindent \textbf{Role of Anatomical Consistency Loss.} Adopting the anatomical consistency loss $\mathcal{L}_{\rm cons.}$ to the baseline also produces a notable gain (\textit{e.g.}, $+0.88\%$ DSC on CAXF).\begin{table}[htbp]
\centering
\renewcommand\arraystretch{1.2}
\begin{tabular}{llll}
\toprule
Setting & SS ($\times10^{-2}$)~$\uparrow$ & CHI~$\uparrow$ & DBI~$\downarrow$ \\ \midrule
\textit{w/o} $\mathcal{L}_{\rm cons.}$ & $-4.19$ & $17.11$ & $25.32$ \\
\cellcolor{gray!20}\textit{w/} $\mathcal{L}_{\rm cons.}$ & \cellcolor{gray!20}$\mathbf{0.54}$ & \cellcolor{gray!20}$\mathbf{607.24}$ & \cellcolor{gray!20}$\mathbf{4.03}$ \\ \bottomrule
\multicolumn{4}{l}{\footnotesize \tiny SS: Silhouette Score; CHI: Calinski-Harabasz Index; DBI: Davies-Bouldin Index.}
\end{tabular}
\caption{Results of clustering metrics on XCAD.}
\label{table:clustering}
\end{table}\begin{figure}[htbp]
	\centerline{\includegraphics{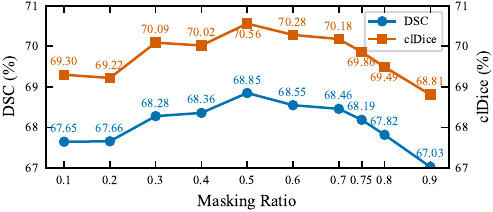}}
	\caption{In-depth analysis of the masking ratio $\gamma$ on ARCADE. $\gamma$ is set to $0.5$ in our default settings.}
	\label{fig:masking_ratio_arcade}
\end{figure}This suggests that $\mathcal{L}_{\rm cons.}$ enables the model to learn more discriminative vascular representations. To further verify the impact of $\mathcal{L}_{\rm cons.}$, we first split patches into vessel‑containing and background-only groups, then use UMAP~\cite{healy2024uniform} to project their representations from the model pre-trained \textit{w/} and \textit{w/o} $\mathcal{L}_{\rm cons.}$ into a low-dimensional space, and evaluate how well these representations cluster. As shown in Table~\ref{table:clustering}, the model pre-trained \textit{w/} $\mathcal{L}_{\rm cons.}$ achieves much higher SS and CHI, and a far lower DBI than the model pre-trained \textit{w/o} $\mathcal{L}_{\rm cons.}$, indicating more compact and well-separated clusters of vascular representations.

\subsection{In-Depth Analysis} We next present a detailed analysis of the proposed VasoMIM. Default settings are highlighted in \sethlcolor{gray!20}\hl{gray}. All results are averaged over three random seeds.

\begin{table}[htbp]
\centering
\renewcommand\arraystretch{1.2}
\begin{tabular}{lllll}
\toprule
Case & $\beta_0$ & $\beta_E$ & ARCADE & CAXF \\ \midrule
Random & $0$ & $0$ & $68.45$ & $84.03$ \\ \midrule
\multirow{3}{*}{Weak-to-Strong} & \cellcolor{gray!20}$0$ & \cellcolor{gray!20}$0.5$ & \cellcolor{gray!20}$\mathbf{68.85}$ & \cellcolor{gray!20}$\mathbf{84.49}$ \\
 & $0$ & $1$ & $68.52$ & $84.24$ \\
 & $1$ & $1$ & $65.36$ & $81.17$ \\ \midrule
 Strong-to-Weak & $0.5$ & $0$ & $67.81$ & $83.41$ \\
\bottomrule
\end{tabular}
\caption{
Effects of different masking strategies. DSC is reported in this table.
}
\label{table:strategy}
\end{table}
\begin{figure}[htbp]
	\centerline{\includegraphics{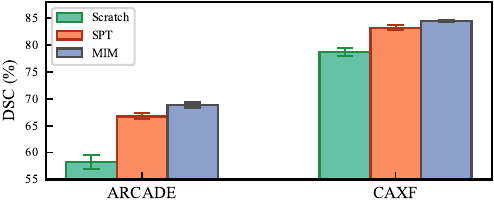}}
	\caption{SPT \textit{vs.} MIM. Our MIM (\textit{i.e.}, VasoMIM) yields average $+2.11\%$ and $+1.31\%$ DSC improvements on ARCADE and CAXF, respectively.}
	\label{fig:spt_mim}
\end{figure}

\noindent \textbf{Masking Ratio.} As illustrated in Fig.~\ref{fig:masking_ratio_arcade}, we observe that a moderate masking ratio (\textit{i.e.}, $0.5$) yields better performance. This result differs slightly from prior work, \textit{e.g.}, MAE adopts a higher ratio of $\gamma = 0.75$. We hypothesize that this discrepancy stems from the unique characteristics of X-ray angiograms. Unlike natural images, which contain objects of varying sizes, vessels in X-ray angiograms occupy a relatively small portion of images. Using a larger masking ratio tends to mask background-only patches that carry little informative content. This may limit the model’s ability to learn useful vascular representations.

\noindent \textbf{Different Masking Strategies.} We evaluate several masking strategies, with results summarized in Table~\ref{table:strategy}. Simply increasing the degree of anatomical guidance does not consistently yield better performance, suggesting that a degree of randomness in the masking process is essential. Specifically, the configuration $\beta_0 = 0$, $\beta_E = 0.5$ achieves the best results, striking a balance between stronger anatomical guidance ($\beta_0 = \beta_E = 1$) and greater randomness ($\beta_0 = \beta_E = 0$). This outcome is quite intuitive. Aggressively masking patches with the highest anatomical relevance leads to most vessel-containing patches being masked. In such cases, the model is forced to reconstruct vascular  anatomy solely from background-only patches with little to no semantic cues.


\noindent \textbf{Strong-to-Weak Anatomical Guidance.} We further investigate a reversed approach, adopting a strong-to-weak strategy during pre-training. As shown in Table~\ref{table:strategy}, this approach leads to a clear performance drop, even compared to using a random masking strategy. This finding underscores the importance of our weak-to-strong masking design.

\begin{figure}[htbp]
	\centerline{\includegraphics{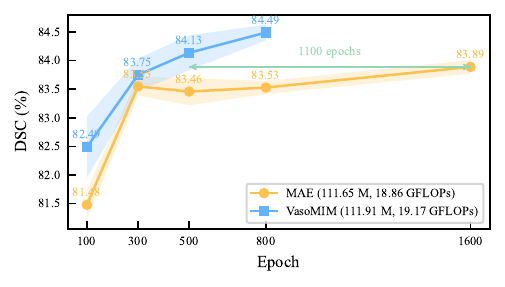}}
	\caption{Fine-tuning on CAXF with models pre-trained for various epochs. GFLOPs are measured on an NVIDIA ${\rm A6000}$ GPU using a single $224\times224$ masked RGB image.}
	\label{fig:epoch_efficiency}
\end{figure}
\begin{table}[htbp]
\centering
\renewcommand\arraystretch{1.2}

\begin{tabular}{lccc}
\toprule
\multirow{2}{*}{Method} & \multicolumn{3}{c}{Ratio (\%)} \\ \cmidrule{2-4} 
 & $25$ & $50$ & $100$ \\ \midrule
Scratch & $38.09$ & $50.67$ & $58.27$ \\
MAE & $56.58$ & $63.04$ & $68.17$ \\
VasoMIM & $\mathbf{56.93}$ & $\mathbf{63.64}$ & $\mathbf{68.85}$ \\ \bottomrule
\end{tabular}
\caption{
Fine-tuning with $25\%$, $50\%$, and $100\%$ training data on ARCADE. DSC is reported in this table.
}
\label{table:data_efficiency}
\end{table}

\noindent \textbf{Segmentor-based Pre-Training \textit{vs.} MIM.} We compare a segmentor-based pre-training (SPT) approach with our MIM using the same segmentor (\textit{i.e.}, U-Net). In SPT, the segmentor is pre-trained in a fully supervised manner using pseudo-labels generated by Frangi filter on the pre-training dataset and subsequently fine-tuned on downstream datasets. Fig.~\ref{fig:spt_mim} demonstrates that both pre-training paradigms substantially outperform the model trained from scratch. Moreover, MIM further boosts DSC from $66.74\%$ to $68.85\%$ ($+2.11\%$) on ARCADE and from $83.18\%$ to $84.49\%$ ($+1.31\%$) on CAXF. Overall, these results indicate that our MIM learns more robust and transferable representations than SPT.

\noindent \textbf{Pre-Training and Data Efficiency.} VasoMIM’s advantages become even more pronounced under resource-limited scenarios. First, we fine-tune MAE and VasoMIM pre-trained for $100$, $300$, $500$, and $800$ epochs on CAXF. As shown in Fig.~\ref{fig:epoch_efficiency}, VasoMIM consistently outperforms MAE at every pre-training duration. Notably, after only $300$ epochs of pre-training ($\sim6$ hours), VasoMIM achieves a DSC of $83.75\%$ on CAXF, slightly above MAE’s $83.53\%$, which requires $800$ epochs ($\sim12$ hours). Even when we extend MAE’s pre-training to $1,600$ epochs ($\sim24$ hours) for a further performance boost, VasoMIM pre-trained for only 500 epochs ($\sim10$ hours) still performs better. Next, to test data efficiency, we fine-tune models on ARCADE using only $25\%$, $50\%$, or $100\%$ of the training labels. Table~\ref{table:data_efficiency} shows that VasoMIM outperforms MAE in all scenarios.

\section{Conclusion}
In this paper, we propose VasoMIM, a vascular anatomy-aware MIM framework specifically tailored for X-ray angiograms. By incorporating anatomical guidance into the masking strategy, our model can focus on vessel-relevant regions. Additionally, the proposed anatomical consistency loss significantly enhances the discriminability of learned vascular representations. Extensive experiments indicate that the proposed framework yields superior performance on three benchmarks. This work provides new perspectives on integrating anatomical knowledge into SSL frameworks.

\section{Acknowledgments}
This work was supported in part by the National Key Research and Development Program of China under Grant 2023YFC2415100, in part by the National Natural Science Foundation of China under Grant 62222316, Grant 62373351, Grant 82327801, Grant 62303463, in part by the Chinese Academy of Sciences Project for Young Scientists in Basic Research under Grant No. YSBR-104, in part by the Beijing Natural Science Foundation under Grant F252068, Grant 4254107, in part by Beijing Nova Program under Grant 20250484813, in part by China Postdoctoral Science Foundation under Grant 2024M763535, in part by the Postdoctoral Fellowship Program of CPSF under Grant GZC20251170.

\section{Supplementary Material}
In this Supplementary Material, we first present detailed descriptions of the datasets used for pre-training, vessel segmentation, and image denoising. Next, we introduce more implementation details. Finally, we provide more quantitative and qualitative results.

\subsection{Dataset Description}
\textbf{Pre-Training.} \textbf{(1) ARCADE}~\cite{popov2024dataset}: This dataset comprises two subsets for coronary vessel segmentation and atherosclerotic plaque localization. Each subset contains $1,500$ images, split into $1,000$ for training, $200$ for validation and $300$ for testing. Only the $1,000$ training images per subset are used for pre-training. \textbf{(2) CADICA}~\cite{jimenez2024cadica}: It comprises coronary X-ray angiography videos from $42$ patients, with frames of $512\times512$ pixels and video lengths ranging from $1$ to $151$ frames. We select $6,594$ high-quality frames from these videos for pre-training. \textbf{(3) Stenosis}~\cite{danilov2021real}: It offers X-ray angiography videos from $100$ patients with a total of $8,325$ frames. We use the $7,492$ training frames (from $90$ patients) for pre-training. \textbf{(4) SYNTAX}~\cite{mahmoudi2025x}: This dataset includes $2,943$ X-ray angiograms from $231$ patients. All images are used for pre-training. \textbf{(5) XCAD}~\cite{ma2021self}: This dataset provides $1,747$ X-ray angiograms, of which $1,621$ are used for pre-training.

\noindent \textbf{Vessel Segmentation.} \textbf{(1) ARCADE}: We fine-tune models on the validation set ($200$ images) of the coronary vessel segmentation subset and report performance on its test set ($300$ images). \textbf{(2) XCAV}~\cite{wu2025denver}: This dataset comprises 111 angiography videos from $59$ patients. We extract frames with pixel-level vessel annotations, splitting them by patient into $175$ training frames ($47$ patients) and $46$ test frames ($12$ patients). \textbf{(3) CAXF}~\cite{li2020cau}: It consists of $538$ images from $36$ X-ray angiography videos. Following the original protocol~\cite{li2020cau}, $337$ frames from $24$ videos are utilized for training, and the remaining $201$ images from $12$ videos are selected for testing.

\begin{table}[htbp]
  \renewcommand{\thetable}{S1}%
  \centering
  \renewcommand\arraystretch{1.2}
  \begin{tabular}{ll}
    \toprule
     Config & Value \\ \midrule
     Optimizer & AdamW \\
     Base learning rate & $1.5e^{-4}$ \\
     Weight decay & $0.05$ \\
     Momentum & $\beta_1,\beta_2=0.9,0.95$ \\
     Layer-wise lr decay & $1.0$ \\
     Batch size& $256$ \\
     Learning rate schedule & Cosine decay \\
     Warmup epochs & $40$ \\
     Training epochs & $800$ \\
     Augmentation & RandomResizedCrop \\ \bottomrule
  \end{tabular}
  \addtocounter{table}{-1}
  \caption{Pre-training configurations of VasoMIM.} \label{table:pre_config}
\end{table}
\begin{table}[htbp]
  \renewcommand{\thetable}{S2}%
  \centering
  \renewcommand\arraystretch{1.2}
  \resizebox{\linewidth}{!}{
  \begin{tabular}{lcc}
    \toprule
    Method & PSNR (dB) & SSIM \\ \midrule
    \multicolumn{3}{l}{\textit{Traditional}} \\
    LSF~\cite{buades2005non} & $28.36$ & $0.6300$ \\ \midrule
    \multicolumn{3}{l}{\textit{From Scratch}} \\
    DnCNN~\cite{zhang2017beyond} & $35.00$ & $0.8822$ \\ \midrule
    \multicolumn{3}{l}{\textit{Self-Supervised Pre-Training}} \\
    DINO~\cite{caron2021emerging} & $35.08$ & \underline{$0.8840$} \\
    MAE~\cite{he2022masked} & \underline{$35.11$} & $0.8837$ \\
    DeblurringMIM$^\dag$~\cite{kang2024deblurring} & $35.10$ & $0.8830$ \\
    HPM~\cite{wang2025bootstrap} & $35.10$ & $0.8835$ \\
    \textbf{VasoMIM} & $\bm{35.14}$ & $\bm{0.8842}$ \\ \bottomrule
\end{tabular}}
  \addtocounter{table}{-1}
  \caption{Comparison of state-of-the-art methods on CoronaryDominance. The best results are highlighted in \textbf{bold} and the second-best results are \underline{underlined}. Results are averaged over three random seeds.} \label{table:image_denoising}
\end{table}

\noindent \textbf{Image Denoising.} X-ray angiography systems often use low-dose protocols to minimize patient radiation exposure~\cite{harbron2016patient}, which introduces substantial noise. Image denoising algorithms can clarify vascular anatomy by suppressing this noise \cite{luo2020ultra}. Since no public dataset exists, we construct one from a subset of \textbf{CoronaryDominance} \cite{kruzhilov2025coronarydominance}. From angiography videos of $249$ patients, we manually select one high-quality frame per video, yielding $1,595$ frames. We then simulate noise by applying Poisson noise, Gaussian noise and Gaussian blur. The resulting dataset is split into $1,286$ training frames ($200$ patients) and $309$ test frames ($49$ patients). For evaluation, we adopt peak signal-to-noise ratio (PSNR) and structural similarity index measure (SSIM) as metrics.

\begin{table}[htbp]
  \renewcommand{\thetable}{S3}%
  \centering
  \renewcommand\arraystretch{1.2}
  \begin{tabular}{llll}
    \toprule
    \multirow{2}{*}{$\alpha$} & \multirow{2}{*}{Method} & \multicolumn{2}{c}{DSC (\%)} \\ \cmidrule{3-4} 
     & & ARCADE & CAXF \\ \midrule
    \multirow{2}{*}{$50$} & Frangi & $12.06$ & $24.46$ \\
     & VasoMIM & $67.97$ & $84.06$ \\ \midrule
    \multirow{2}{*}{$80$} & Frangi & $27.06$ & $25.10$ \\
     & VasoMIM & $68.30$ & $84.29$ \\ \midrule
    \multirow{2}{*}{$92$} & Frangi & $41.30$ & $64.01$ \\
     & \cellcolor{gray!20}VasoMIM & \cellcolor{gray!20}$\mathbf{68.85}$ & \cellcolor{gray!20}$\mathbf{84.49}$ \\ 
    \bottomrule
    \end{tabular}
  \addtocounter{table}{-1}
  \caption{Ablations across varying levels of vascular anatomy quality. Higher $\alpha$ denotes improved vascular anatomy quality.} \label{table:frangi}
\end{table}
\renewcommand{\thefigure}{S1}  
\begin{figure}[htbp]
	\centerline{\includegraphics{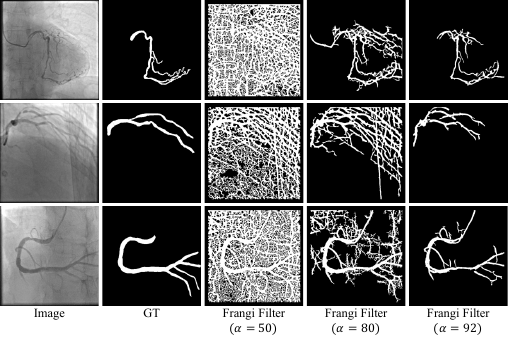}}
	\caption{Qualitative results of Frangi filter under different thresholds $\alpha$.}
	\label{fig:frangi}
\end{figure}
\renewcommand{\thefigure}{\arabic{figure}}  

\subsection{Implementation Details}
\textbf{Pre-Training.} For all experiments, we use the dataset in Table~1, which comprises $20$K X-ray angiograms for pre-training. Most of the configurations are borrowed from MAE~\cite{he2022masked}. By default, we use ViT-B/$16$~\cite{dosovitskiy2021image} as the backbone and pre-train it for $800$ epochs. For the segmentor, we employ the lightweight UNeXt-S~\cite{valanarasu2022unext}, which has only $0.3$ M parameters. We adopt the linear learning rate scaling rule~\cite{goyal2017accurate}: $lr=lr_{\rm base}\times{\rm batch\_size}/256$. Additional implementation details are provided in Table~\ref{table:pre_config}.

\noindent \textbf{Vessel Segmentation.} U-Net~\cite{ronneberger2015u} is used as the segmentation decoder and fine-tuned for $500$ epochs on each dataset. All input images are resized to $224\times 224$. We employ the AdamW optimizer with an initial learning rate of $1e^{-4}$ and a weight decay of $0.05$. A cosine-annealing learning rate schedule with $T\_\max=500$ is applied.

\subsection{More Results}
\textbf{Quantitative Results on Image Denoising.} As shown in Table~\ref{table:image_denoising}, VasoMIM attains a leading PSNR of $35.14$ dB and SSIM of $0.8842$, outperforming all baseline methods. This consistent gain underscores the superior transferability of the vascular representations learned by our framework.

\renewcommand{\thefigure}{S2}  
\begin{figure}[!htbp]
	\centerline{\includegraphics{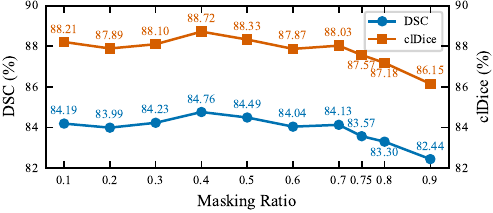}}
	\caption{Effects of the masking ratio on CAXF. $\gamma$ is set to $0.5$ in our default settings.}
	\label{fig:masking_ratio_caxf}
\end{figure}
\renewcommand{\thefigure}{\arabic{figure}}  
\begin{table}[!htbp]
  \renewcommand{\thetable}{S4}%
  \centering
  \renewcommand\arraystretch{1.2}
  \begin{tabular}{lll}
    \toprule
    \multirow{2}{*}{Setting} & \multicolumn{2}{c}{DSC (\%)} \\ 
    \cmidrule(lr){2-3}
     & ARCADE & CAXF \\ 
    \midrule
    \textit{w/o} $\mathcal{L}_{\rm cons.}$ & $68.30$ & $83.96$ \\
    Joint
      & $68.78$ 
      & $84.06$ \\
    \cellcolor{gray!20}In Advance 
      & \cellcolor{gray!20}$\mathbf{68.85}$ & \cellcolor{gray!20}$\mathbf{84.49}$ \\ 
    \bottomrule
  \end{tabular}
  \addtocounter{table}{-1}
  \caption{Comparison of different training settings for the segmentor $\mathcal{S}$ used in the anatomical consistency loss, Eq.~(4).} \label{table:advance_jointly}
\end{table}
\begin{table}[!htbp]
  \renewcommand{\thetable}{S5}%
  \centering
  \renewcommand\arraystretch{1.2}
  \begin{tabular}{lll}
    \toprule
    \multirow{2}{*}{Loss} & \multicolumn{2}{c}{DSC (\%)} \\ 
    \cmidrule(lr){2-3}
     & ARCADE & CAXF \\ 
    \midrule
    -- & $68.30$ & $83.96$ \\
    $\mathcal{L}_{\rm w-rec.}$
      & $68.78$ & $84.14$ \\
    \cellcolor{gray!20}$\mathcal{L}_{\rm cons.}$ 
      & \cellcolor{gray!20}$\mathbf{68.85}$ & \cellcolor{gray!20}$\mathbf{84.49}$ \\ 
    \bottomrule
  \end{tabular}
  \addtocounter{table}{-1}
  \caption{Anatomy-weighted reconstruction loss $\mathcal{L}_{\rm w-rec.}$ \textit{vs.} Anatomy consistency loss $\mathcal{L}_{\rm cons.}$.} \label{table:weighted_loss}
\end{table}

\noindent \textbf{Effects of Vascular Anatomy Quality.} The threshold $\alpha$ within Frangi filter~\cite{frangi1998multiscale} controls the quality of the extracted vascular anatomy, as shown in Fig.~\ref{fig:frangi}. As $\alpha$ increases, the performance of Frangi filter improves significantly, while VasoMIM’s performance only rises marginally, as shown in Table~\ref{table:frangi}. This demonstrates that our anatomy-aware pre-training framework is highly robust to the quality of vascular anatomy.

\noindent \textbf{Influence of Masking Ratio on CAXF.} To further investigate the sensitivity to the masking ratio on CAXF, we vary $\gamma$ and present the results in Fig.~\ref{fig:masking_ratio_caxf}. Consistent with ARCADE, VasoMIM performs best at intermediate ratios ($0.3\sim0.5$), so we adopt $\gamma=0.5$ by default.

\noindent \textbf{Training in Advance \textit{vs.} Joint Training.} In VasoMIM, we train the segmentor in advance using pseudo-labels produced by Frangi filter and keep its weights frozen during MIM pre-training. We also evaluate a joint training variant, where the segmentor and the MIM model are optimized simultaneously during pre-training. As Table \ref{table:advance_jointly} shows, training in advance yields larger DSC gains than joint training, while joint training increases pre-training time from $15.5$ to $20.5$ hours. Thus, training the segmentor in advance is both more effective and more efficient.

\begin{table}[htbp]
  \renewcommand{\thetable}{S6}%
  \centering
  \renewcommand\arraystretch{1.2}
  \begin{tabular}{lll}
    \toprule
    \multirow{2}{*}{$\mathcal{L}$} & \multicolumn{2}{c}{DSC (\%)} \\ 
    \cmidrule(lr){2-3}
     & ARCADE & CAXF \\ 
    \midrule
    -- & $68.30$ & $83.96$ \\
    $\text{L}_1$ 
      & $68.81$ & $84.32$ \\
    Dice 
      & $67.97$ & $84.12$ \\
    \cellcolor{gray!20}Cross Entropy 
      & \cellcolor{gray!20}$\mathbf{68.85}$ & \cellcolor{gray!20}$\mathbf{84.49}$  \\ 
    \bottomrule
  \end{tabular}
  \addtocounter{table}{-1}
  \caption{Sensitivity to different metric functions $\mathcal{L}$.} \label{table:loss}
\end{table}
\begin{table}[!htbp]
  \renewcommand{\thetable}{S7}%
  \centering
  \renewcommand\arraystretch{1.2}
  \begin{tabular}{llll}
    \toprule
    \multirow{2}{*}{Blocks} & \multirow{2}{*}{\# params (M)} & \multicolumn{2}{c}{DSC (\%)} \\ \cmidrule{3-4} 
     & & ARCADE & CAXF \\ \midrule
    $1$ & $89.84~(1.25\times)$ & $\mathbf{68.86}$ & $84.38$ \\
    $2$ & $92.99~(1.20\times)$ & $68.39$ & $84.22$ \\
    $4$ & $99.30~(1.13\times)$ & $68.11$ & $83.85$ \\
    \cellcolor{gray!20}$8$ & \cellcolor{gray!20}$111.91~(1.00\times)$ & \cellcolor{gray!20}$68.85$ & \cellcolor{gray!20}$\mathbf{84.49}$ \\
    $12$ & $124.52~(0.90\times)$ & $68.75$ & $84.31$ \\
    \midrule \midrule
    \multirow{2}{*}{Dim} & \multirow{2}{*}{\# params (M)} & \multicolumn{2}{c}{DSC (\%)} \\ \cmidrule{3-4} 
     & & ARCADE & CAXF \\ \midrule
    $128$ & $87.68~(1.28\times)$ & $67.58$ & $83.59$ \\
    $256$ & $92.61~(1.21\times)$ & $66.68$ & $82.47$ \\ 
    \cellcolor{gray!20}$512$ & \cellcolor{gray!20}$111.91~(1.00\times)$ & \cellcolor{gray!20}$\mathbf{68.85}$ & \cellcolor{gray!20}$\mathbf{84.49}$ \\ 
    $1024$ & $188.25~(0.59\times)$ & $68.16$ & $84.23$ \\
    \bottomrule
  \end{tabular}
  \addtocounter{table}{-1}
  \caption{Sensitivity to different decoder configurations.} \label{table:decoder}
\end{table}

\noindent \textbf{Anatomy-Weighted Reconstruction Loss \textit{vs.} Anatomical Consistency Loss.} In this section, we compare our anatomical consistency loss $\mathcal{L}_{\rm cons.}$ against an anatomy-weighted reconstruction loss $\mathcal{L}_{\rm w-rec.}$, where the standard reconstruction loss $\mathcal{L}_{\rm rec.}$ is weighted by the patch-wise vessel distribution $f$. The overall training loss of this variant is $\mathcal{L}_{\rm train}^\prime=\mathcal{L}_{\rm rec.}+\mathcal{L}_{\rm w-rec.}$. As Table~\ref{table:weighted_loss} shows, $\mathcal{L}_{\rm w-rec.}$ improves DSC by $0.48\%$ on ARCADE and $0.18\%$ on CAXF, while $\mathcal{L}_{\rm cons.}$ yields larger gains of $0.55\%$ and $0.53\%$, respectively. These results show that $\mathcal{L}_{\rm cons.}$ enables the model to learn more robust vascular representations.

\noindent \textbf{Different Metric Functions $\mathcal{L}$.} We study different metric functions in Eq.~(4), including $\text{L}_1$ loss, Dice loss, and CE loss. As shown in Table~\ref{table:loss}, we find that CE loss is the best choice, improving DSC by $0.55\%$ on ARCADE and $0.53\%$ on CAXF compared to the baseline.

\noindent \textbf{Different Decoder Designs.} Following~\cite{he2022masked}, we investigate two key decoder design choices: depth and embedding dimension. The results are summarized in Table~\ref{table:decoder}. Consistent with MAE and its variants (Table~2), the configuration with $8$ decoder blocks and an embedding dimension of $512$ is the best choice.

\begin{table}[!htbp]
  \renewcommand{\thetable}{S8}%
  \centering
  \renewcommand\arraystretch{1.2}
  \resizebox{\linewidth}{!}{
    \begin{tabular}{llll}
    \toprule
    Method & \# params (M) & GFLOPs & Speed (s/epoch) \\ \midrule
    MAE & $111.65~(1.00\times)$ & $18.86~(1.02\times)$ & $54~(1.30\times)$ \\
    VasoMIM & $111.91~(1.00\times)$ & $19.17~(1.00\times)$ & $70~(1.00\times)$ \\ \bottomrule
    \end{tabular}
    }
  \addtocounter{table}{-1}
  \caption{Model Complexity. GFLOPs and speed are evaluated on a single NVIDIA ${\rm A6000}$ GPU. GFLOPs are computed using a single $224\times 224$ RGB image (with masking) as the input. Speed is evaluated with a batch size of $256$.} \label{table:complexity}
\end{table}

\noindent \textbf{Model Complexity.} We compare the model complexity of MAE and our VasoMIM in Table~\ref{table:complexity}. Integrating UNeXt‑S~\cite{valanarasu2022unext} into VasoMIM adds only negligible overhead. Considering the training efficiency gains reported in Fig.~7, VasoMIM is the more efficient choice overall.

\noindent \textbf{Qualitative Results on Vessel Segmentation.} We present additional qualitative results on ARCADE, CAXF, and XCAV. As shown in Fig.~\ref{fig:sota_vis_appendix}, VasoMIM achieves more precise vessel segmentation results with fewer false positives.

\noindent \textbf{Visualization of Patch-Wise Masking Ratio.} We visualize additional examples of patch-wise masking ratio over the pre-training process, clearly demonstrating that VasoMIM preferentially masks patches containing vessels.

\renewcommand{\thefigure}{S3}  
\begin{figure*}[t]
	\centering
	\includegraphics{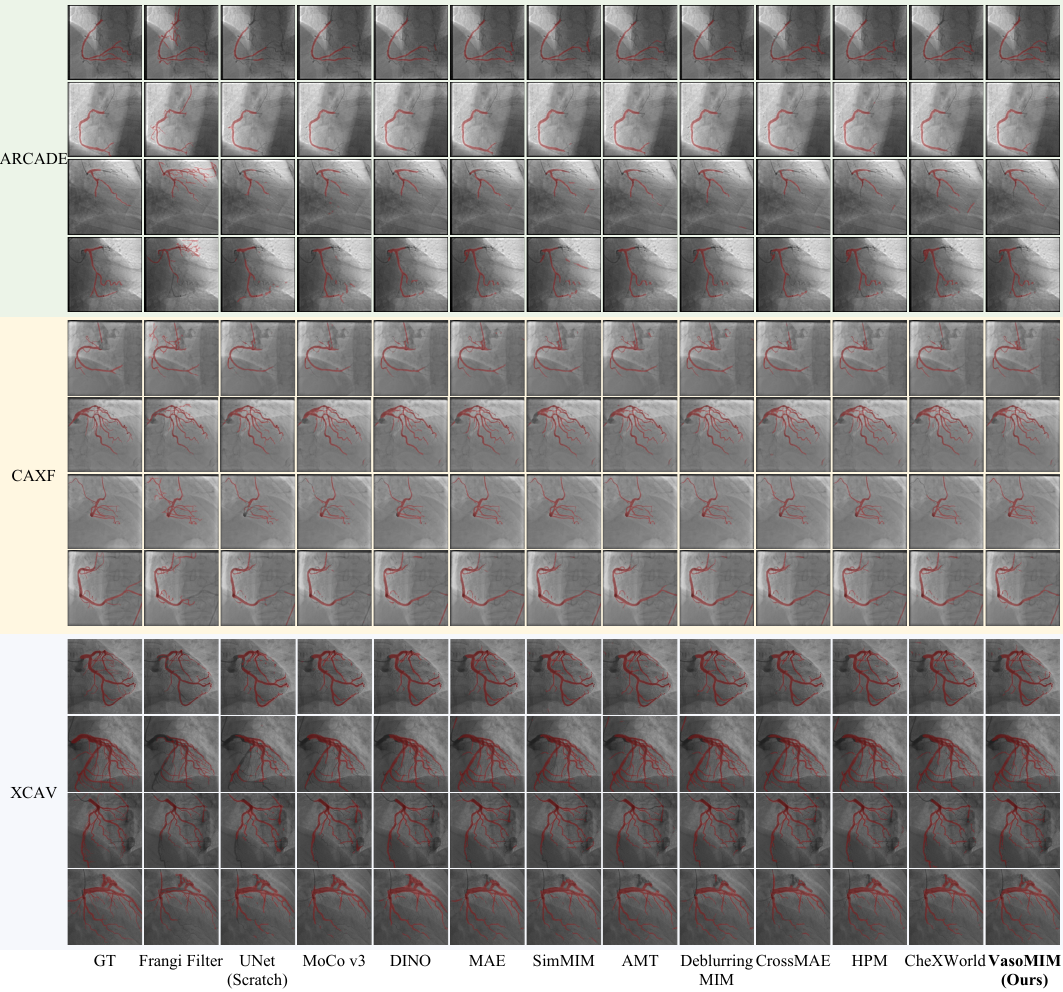}
	\caption{More qualitative results on ARCADE, CAXF, and XCAV.}
	\label{fig:sota_vis_appendix}
\end{figure*}
\renewcommand{\thefigure}{\arabic{figure}}  
\renewcommand{\thefigure}{S4}  
\begin{figure*}[t]
	\centering
	\centerline{\includegraphics{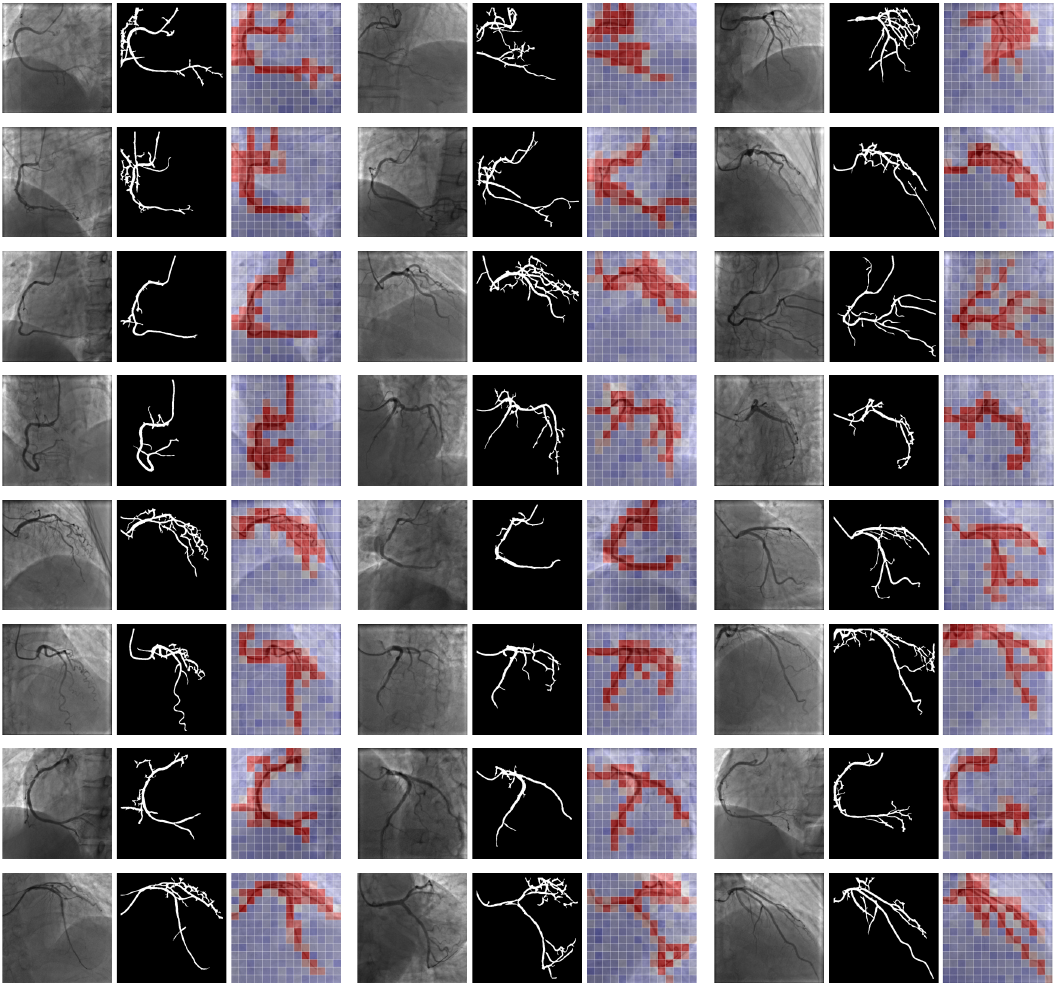}}
	\caption{Patch-wise masking ratio over the pre-training process, \textit{i.e.}, $\frac{1}{E}\sum_{j=1}^E\mathbb{I}\left(\text{Patch $x_i$ is masked in epoch $j$}\right)$. For each case, we show the \textit{input image}, \textit{vascular anatomy} extracted by Frangi filter, and \textit{patch-wise masking ratio} over pre-training. Red means higher ratio and blue indicates the opposite.}
	\label{fig:ratio_vis_appendix}
\end{figure*}
\renewcommand{\thefigure}{\arabic{figure}}  

\clearpage

\bibliography{aaai2026}

\begin{thebibliography}{63}
\providecommand{\natexlab}[1]{#1}

\bibitem[{Adams and Bischof(1994)}]{adams1994seeded}
Adams, R.; and Bischof, L. 1994.
\newblock Seeded region growing.
\newblock \emph{IEEE TPAMI}, 16(6): 641--647.

\bibitem[{Azizi et~al.(2021)}]{azizi2021big}
Azizi, S.; et~al. 2021.
\newblock Big self-supervised models advance medical image classification.
\newblock In \emph{Proc. CVPR}, 3478--3488.

\bibitem[{Bao et~al.(2022)Bao, Dong, Piao, and Wei}]{bao2022beit}
Bao, H.; Dong, L.; Piao, S.; and Wei, F. 2022.
\newblock {BEIT}: {BERT} pre-training of image transformers.
\newblock In \emph{Proc. ICLR}.

\bibitem[{Buades, Coll, and Morel(2005)}]{buades2005non}
Buades, A.; Coll, B.; and Morel, J.-M. 2005.
\newblock A non-local algorithm for image denoising.
\newblock In \emph{Proc. CVPR}, volume~2, 60--65.

\bibitem[{Caron et~al.(2021)}]{caron2021emerging}
Caron, M.; et~al. 2021.
\newblock Emerging properties in self-supervised vision transformers.
\newblock In \emph{Proc. ICCV}, 9650--9660.

\bibitem[{Chaitanya et~al.(2020)Chaitanya, Erdil, Karani, and Konukoglu}]{chaitanya2020contrastive}
Chaitanya, K.; Erdil, E.; Karani, N.; and Konukoglu, E. 2020.
\newblock Contrastive learning of global and local features for medical image segmentation with limited annotations.
\newblock In \emph{Proc. NeurIPS}, volume~33, 12546--12558.

\bibitem[{Chen et~al.(2024)}]{chen2024transunet}
Chen, J.; et~al. 2024.
\newblock {TransUNet}: Rethinking the {U-Net} architecture design for medical image segmentation through the lens of transformers.
\newblock \emph{MedIA}, 97: 103280.

\bibitem[{Chen et~al.(2020)Chen, Kornblith, Norouzi, and Hinton}]{chen2020simple}
Chen, T.; Kornblith, S.; Norouzi, M.; and Hinton, G. 2020.
\newblock A simple framework for contrastive learning of visual representations.
\newblock In \emph{Proc. ICML}, 1597--1607.

\bibitem[{Chen, Xie, and He(2021)}]{chen2021empirical}
Chen, X.; Xie, S.; and He, K. 2021.
\newblock An empirical study of training self-supervised vision transformers.
\newblock In \emph{Proc. CVPR}, 9640--9649.

\bibitem[{Danilov et~al.(2021)}]{danilov2021real}
Danilov, V.~V.; et~al. 2021.
\newblock Real-time coronary artery stenosis detection based on modern neural networks.
\newblock \emph{Sci. Rep.}, 11(1): 7582.

\bibitem[{Dosovitskiy et~al.(2021)}]{dosovitskiy2021image}
Dosovitskiy, A.; et~al. 2021.
\newblock An image is worth 16x16 words: Transformers for image recognition at scale.
\newblock In \emph{Proc. ICLR}.

\bibitem[{Esteva et~al.(2019)}]{esteva2019guide}
Esteva, A.; et~al. 2019.
\newblock A guide to deep learning in healthcare.
\newblock \emph{Nat. Med.}, 25(1): 24--29.

\bibitem[{Frangi et~al.(1998)Frangi, Niessen, Vincken, and Viergever}]{frangi1998multiscale}
Frangi, A.~F.; Niessen, W.~J.; Vincken, K.~L.; and Viergever, M.~A. 1998.
\newblock Multiscale vessel enhancement filtering.
\newblock In \emph{Proc. MICCAI}, 130--137.

\bibitem[{Fu et~al.(2025)}]{fu2025rethinking}
Fu, L.; et~al. 2025.
\newblock Rethinking patch dependence for masked autoencoders.
\newblock \emph{TMLR}.

\bibitem[{Goyal et~al.(2017)}]{goyal2017accurate}
Goyal, P.; et~al. 2017.
\newblock Accurate, large minibatch {SGD}: Training {ImageNet} in 1 hour.
\newblock \emph{arXiv}.

\bibitem[{Gui et~al.(2024)}]{gui2024survey}
Gui, J.; et~al. 2024.
\newblock A survey on self-supervised learning: Algorithms, applications, and future trends.
\newblock \emph{IEEE TPAMI}, 46(12): 9052--9071.

\bibitem[{Harbron et~al.(2016)}]{harbron2016patient}
Harbron, R.; et~al. 2016.
\newblock Patient radiation doses in paediatric interventional cardiology procedures: A review.
\newblock \emph{J. Radiol. Prot.}, 36(4): R131.

\bibitem[{Hatamizadeh et~al.(2021)Hatamizadeh, Nath, Tang, Yang, Roth, and Xu}]{hatamizadeh2021swin}
Hatamizadeh, A.; Nath, V.; Tang, Y.; Yang, D.; Roth, H.~R.; and Xu, D. 2021.
\newblock Swin {UNETR}: {Swin} transformers for semantic segmentation of brain tumors in {MRI} images.
\newblock In \emph{Proc. MICCAIW}, 272--284.

\bibitem[{He et~al.(2022)He, Chen, Xie, Li, Doll{\'a}r, and Girshick}]{he2022masked}
He, K.; Chen, X.; Xie, S.; Li, Y.; Doll{\'a}r, P.; and Girshick, R. 2022.
\newblock Masked autoencoders are scalable vision learners.
\newblock In \emph{Proc. CVPR}, 16000--16009.

\bibitem[{Healy and McInnes(2024)}]{healy2024uniform}
Healy, J.; and McInnes, L. 2024.
\newblock Uniform manifold approximation and projection.
\newblock \emph{Nat. Rev. Methods Primers}, 4(1): 82.

\bibitem[{Hinojosa, Liu, and Ghanem(2024)}]{hinojosa2024colormae}
Hinojosa, C.; Liu, S.; and Ghanem, B. 2024.
\newblock {ColorMAE}: Exploring data-independent masking strategies in masked autoencoders.
\newblock In \emph{Proc. ECCV}, 432--449.

\bibitem[{Huang et~al.(2024)}]{huang2024spironet}
Huang, D.-X.; et~al. 2024.
\newblock {SPIRONet}: Spatial-frequency learning and topological channel interaction network for vessel segmentation.
\newblock \emph{arXiv}.

\bibitem[{Huang et~al.(2025)}]{huang2025real}
Huang, D.-X.; et~al. 2025.
\newblock Real-time 2{D}/3{D} registration via {CNN} regression and centroid alignment.
\newblock \emph{IEEE TASE}, 22: 85--98.

\bibitem[{Jim{\'e}nez-Partinen et~al.(2024)}]{jimenez2024cadica}
Jim{\'e}nez-Partinen, A.; et~al. 2024.
\newblock {CADICA}: A new dataset for coronary artery disease detection by using invasive coronary angiography.
\newblock \emph{Expert Syst.}, 41(12): e13708.

\bibitem[{Kakogeorgiou et~al.(2022)}]{kakogeorgiou2022hide}
Kakogeorgiou, I.; et~al. 2022.
\newblock What to hide from your students: Attention-guided masked image modeling.
\newblock In \emph{Proc. ECCV}, 300--318.

\bibitem[{Kang et~al.(2024)}]{kang2024deblurring}
Kang, Q.; et~al. 2024.
\newblock Deblurring masked image modeling for ultrasound image analysis.
\newblock \emph{MedIA}, 97: 103256.

\bibitem[{Kheiri et~al.(2022)Kheiri, Simpson, Osman, German, Fuss, and Ferencik}]{kheiri2022computed}
Kheiri, B.; Simpson, T.~F.; Osman, M.; German, D.~M.; Fuss, C.~S.; and Ferencik, M. 2022.
\newblock Computed tomography vs invasive coronary angiography in patients with suspected coronary artery disease: A meta-analysis.
\newblock \emph{JACC: Cardiovasc. Imaging}, 15(12): 2147--2149.

\bibitem[{Kruzhilov et~al.(2025)}]{kruzhilov2025coronarydominance}
Kruzhilov, I.; et~al. 2025.
\newblock {CoronaryDominance}: Angiogram dataset for coronary dominance classification.
\newblock \emph{Sci. Data}, 12(1): 341.

\bibitem[{Li et~al.(2022{\natexlab{a}})Li, Zheng, Liu, Wang, Su, and Zheng}]{li2022semmae}
Li, G.; Zheng, H.; Liu, D.; Wang, C.; Su, B.; and Zheng, C. 2022{\natexlab{a}}.
\newblock {SemMAE}: Semantic-guided masking for learning masked autoencoders.
\newblock In \emph{Proc. NeurIPS}, volume~35, 14290--14302.

\bibitem[{Li et~al.(2020)Li, Bian, Zhou, Xie, Ni, and Hou}]{li2020cau}
Li, R.-Q.; Bian, G.-B.; Zhou, X.-H.; Xie, X.; Ni, Z.-L.; and Hou, Z. 2020.
\newblock {CAU}-net: A novel convolutional neural network for coronary artery segmentation in digital substraction angiography.
\newblock In \emph{Proc. ICONIP}, 185--196.

\bibitem[{Li et~al.(2022{\natexlab{b}})Li, Wang, Yang, and Yang}]{li2022uniform}
Li, X.; Wang, W.; Yang, L.; and Yang, J. 2022{\natexlab{b}}.
\newblock Uniform masking: Enabling {MAE} pre-training for pyramid-based vision transformers with locality.
\newblock \emph{arXiv}.

\bibitem[{Li et~al.(2024)Li, Luan, Wu, Pan, Chen, and Yang}]{li2024anatomask}
Li, Y.; Luan, T.; Wu, Y.; Pan, S.; Chen, Y.; and Yang, X. 2024.
\newblock {AnatoMask}: Enhancing medical image segmentation with reconstruction-guided self-masking.
\newblock In \emph{Proc. ECCV}, 146--163.

\bibitem[{Liu, Gui, and Luo(2023)}]{liu2023good}
Liu, Z.; Gui, J.; and Luo, H. 2023.
\newblock Good helper is around you: Attention-driven masked image modeling.
\newblock In \emph{Proc. AAAI}, volume~37, 1799--1807.

\bibitem[{Loshchilov and Hutter(2019)}]{loshchilov2019decoupled}
Loshchilov, I.; and Hutter, F. 2019.
\newblock Decoupled weight decay regularization.
\newblock In \emph{Proc. ICLR}.

\bibitem[{Luo et~al.(2020)Luo, Majoe, Kui, Qi, Pushparajah, and Rhode}]{luo2020ultra}
Luo, Y.; Majoe, S.; Kui, J.; Qi, H.; Pushparajah, K.; and Rhode, K. 2020.
\newblock Ultra-dense denoising network: Application to cardiac catheter-based {X}-ray procedures.
\newblock \emph{IEEE TBME}, 68(9): 2626--2636.

\bibitem[{Ma et~al.(2021)}]{ma2021self}
Ma, Y.; et~al. 2021.
\newblock Self-supervised vessel segmentation via adversarial learning.
\newblock In \emph{Proc. ICCV}, 7536--7545.

\bibitem[{Mahmoudi et~al.(2025)}]{mahmoudi2025x}
Mahmoudi, S.~S.; et~al. 2025.
\newblock X-ray coronary angiogram images and {SYNTAX} score to develop machine-learning algorithms for {CHD} diagnosis.
\newblock \emph{Sci. Data}, 12(1): 471.

\bibitem[{Members et~al.(2022)}]{writing20222021}
Members, W.~C.; et~al. 2022.
\newblock 2021 {ACC/AHA/SCAI} guideline for coronary artery revascularization: A report of the {American College of Cardiology/American Heart Association Joint Committee} on clinical practice guidelines.
\newblock \emph{JACC}, 79(2): e21--e129.

\bibitem[{P{\'e}rez-Garc{\'\i}a et~al.(2025)}]{perez2025exploring}
P{\'e}rez-Garc{\'\i}a, F.; et~al. 2025.
\newblock Exploring scalable medical image encoders beyond text supervision.
\newblock \emph{Nat. Mach. Intell.}, 7: 119--130.

\bibitem[{Popov et~al.(2024)}]{popov2024dataset}
Popov, M.; et~al. 2024.
\newblock Dataset for automatic region-based coronary artery disease diagnostics using {X}-ray angiography images.
\newblock \emph{Sci. Data}, 11(1): 20.

\bibitem[{Ronneberger et~al.(2015)}]{ronneberger2015u}
Ronneberger, O.; et~al. 2015.
\newblock {U-Net}: Convolutional networks for biomedical image segmentation.
\newblock In \emph{Proc. MICCAI}, 234--241.

\bibitem[{Ruan, Li, and Xiang(2024)}]{ruan2024vm}
Ruan, J.; Li, J.; and Xiang, S. 2024.
\newblock {VM-UNet}: Vision {M}amba {UNet} for medical image segmentation.
\newblock \emph{arXiv}.

\bibitem[{Shit et~al.(2021)}]{shit2021cldice}
Shit, S.; et~al. 2021.
\newblock {clDice}: A novel topology-preserving loss function for tubular structure segmentation.
\newblock In \emph{Proc. CVPR}, 16560--16569.

\bibitem[{Taghizadeh~Dehkordi et~al.(2014)Taghizadeh~Dehkordi, Doost~Hoseini, Sadri, and Soltanianzadeh}]{taghizadeh2014local}
Taghizadeh~Dehkordi, M.; Doost~Hoseini, A.~M.; Sadri, S.; and Soltanianzadeh, H. 2014.
\newblock Local feature fitting active contour for segmenting vessels in angiograms.
\newblock \emph{IET CV}, 8(3): 161--170.

\bibitem[{Tang et~al.(2025)}]{tang2025mambamim}
Tang, F.; et~al. 2025.
\newblock {MambaMIM}: Pre-training {Mamba} with state space token interpolation and its application to medical image segmentation.
\newblock \emph{MedIA}, 103606.

\bibitem[{Vaduganathan et~al.(2022)Vaduganathan, Mensah, Turco, Fuster, and Roth}]{vaduganathan2022global}
Vaduganathan, M.; Mensah, G.~A.; Turco, J.~V.; Fuster, V.; and Roth, G.~A. 2022.
\newblock The global burden of cardiovascular diseases and risk: A compass for future health.
\newblock \emph{JACC}, 80(25): 2361--2371.

\bibitem[{Valanarasu and Patel(2022)}]{valanarasu2022unext}
Valanarasu, J. M.~J.; and Patel, V.~M. 2022.
\newblock {UNeXt}: {MLP}-based rapid medical image segmentation network.
\newblock In \emph{Proc. MICCAI}, 23--33.

\bibitem[{Wang et~al.(2020)}]{wang2020tensor}
Wang, C.; et~al. 2020.
\newblock Tensor-cut: A tensor-based graph-cut blood vessel segmentation method and its application to renal artery segmentation.
\newblock \emph{MedIA}, 60: 101623.

\bibitem[{Wang et~al.(2025)}]{wang2025bootstrap}
Wang, H.; et~al. 2025.
\newblock Bootstrap masked visual modeling via hard patch mining.
\newblock \emph{IEEE TPAMI}.
\newblock {D}OI:10.1109/TPAMI.2025.3557001.

\bibitem[{Wang et~al.(2024)Wang, Chen, Chen, and Wu}]{wang2024lkm}
Wang, J.; Chen, J.; Chen, D.; and Wu, J. 2024.
\newblock {LKM-UNet}: Large kernel vision {Mamba} {UNet} for medical image segmentation.
\newblock In \emph{Proc. MICCAI}, 360--370.

\bibitem[{Wu et~al.(2025)}]{wu2025denver}
Wu, C.-H.; et~al. 2025.
\newblock {DeNVeR}: Deformable neural vessel representations for unsupervised video vessel segmentation.
\newblock In \emph{Proc. CVPR}, 15682--15692.

\bibitem[{Wu, Zhuang, and Chen(2024)}]{wu2024voco}
Wu, L.; Zhuang, J.; and Chen, H. 2024.
\newblock {VoCo}: A simple-yet-effective volume contrastive learning framework for 3{D} medical image analysis.
\newblock In \emph{Proc. CVPR}, 22873--22882.

\bibitem[{Xie et~al.(2024)Xie, Gu, Harada, Zhang, Xia, and Wu}]{xie2024rethinking}
Xie, Y.; Gu, L.; Harada, T.; Zhang, J.; Xia, Y.; and Wu, Q. 2024.
\newblock Rethinking masked image modelling for medical image representation.
\newblock \emph{MedIA}, 98: 103304.

\bibitem[{Xie et~al.(2022)}]{xie2022simmim}
Xie, Z.; et~al. 2022.
\newblock {SimMIM}: A simple framework for masked image modeling.
\newblock In \emph{Proc. CVPR}, 9653--9663.

\bibitem[{Xu et~al.(2025)Xu, Liu, Xu, and Lukasiewicz}]{xu2025self}
Xu, Z.; Liu, Y.; Xu, G.; and Lukasiewicz, T. 2025.
\newblock Self-supervised medical image segmentation using deep reinforced adaptive masking.
\newblock \emph{IEEE TMI}, 44(1): 180--193.

\bibitem[{Yuan et~al.(2023)}]{yuan2023hap}
Yuan, J.; et~al. 2023.
\newblock {HAP}: Structure-aware masked image modeling for human-centric perception.
\newblock In \emph{Proc. NeurIPS}, volume~36, 50597--50616.

\bibitem[{Yue et~al.(2025)Yue, Wang, Tao, Liu, Song, and Huang}]{yue2025chexworld}
Yue, Y.; Wang, Y.; Tao, C.; Liu, P.; Song, S.; and Huang, G. 2025.
\newblock {CheXWorld}: Exploring image world modeling for radiograph representation learning.
\newblock In \emph{Proc. CVPR}, 20778--20788.

\bibitem[{Zhang et~al.(2017)Zhang, Zuo, Chen, Meng, and Zhang}]{zhang2017beyond}
Zhang, K.; Zuo, W.; Chen, Y.; Meng, D.; and Zhang, L. 2017.
\newblock Beyond a {G}aussian denoiser: Residual learning of deep {CNN} for image denoising.
\newblock \emph{IEEE TIP}, 26(7): 3142--3155.

\bibitem[{Zhou et~al.(2020)Zhou, Yu, Bian, Hu, Ma, and Zheng}]{zhou2020comparing}
Zhou, H.-Y.; Yu, S.; Bian, C.; Hu, Y.; Ma, K.; and Zheng, Y. 2020.
\newblock Comparing to learn: Surpassing imagenet pretraining on radiographs by comparing image representations.
\newblock In \emph{Proc. MICCAI}, 398--407.

\bibitem[{Zhou et~al.(2021)}]{zhou2021review}
Zhou, S.~K.; et~al. 2021.
\newblock A review of deep learning in medical imaging: Imaging traits, technology trends, case studies with progress highlights, and future promises.
\newblock \emph{Proc. IEEE}, 109(5): 820--838.

\bibitem[{Zhou et~al.(2019)Zhou, Siddiquee, Tajbakhsh, and Liang}]{zhou2019unet++}
Zhou, Z.; Siddiquee, M. M.~R.; Tajbakhsh, N.; and Liang, J. 2019.
\newblock {UNet}++: Redesigning skip connections to exploit multiscale features in image segmentation.
\newblock \emph{IEEE TMI}, 39(6): 1856--1867.

\bibitem[{Zhuang et~al.(2025{\natexlab{a}})Zhuang, Luo, Wang, Wu, Luo, and Chen}]{zhuang2025advancing}
Zhuang, J.; Luo, L.; Wang, Q.; Wu, M.; Luo, L.; and Chen, H. 2025{\natexlab{a}}.
\newblock Advancing volumetric medical image segmentation via global-local masked autoencoders.
\newblock \emph{IEEE TMI}.
\newblock {D}OI:10.1109/TMI.2025.3569782.

\bibitem[{Zhuang et~al.(2025{\natexlab{b}})}]{zhuang2025mim}
Zhuang, J.; et~al. 2025{\natexlab{b}}.
\newblock {MiM}: Mask in mask self-supervised pre-training for 3{D} medical image analysis.
\newblock \emph{IEEE TMI}.
\newblock {D}OI:10.1109/TMI.2025.3564382.

\end{thebibliography}

\end{document}